\documentclass[10pt,twocolumn,letterpaper]{article}

\usepackage{cvpr}
\usepackage{times}
\usepackage{epsfig}
\usepackage{graphicx}
\usepackage{amsmath}
\usepackage{amssymb}
\usepackage{pifont}
\usepackage{mdwlist}
\usepackage{color}

\usepackage[breaklinks=true,bookmarks=false]{hyperref}

\cvprfinalcopy 

\def\cvprPaperID{****} 

\begin{document}

\title{SiamMan: Siamese Motion-aware Network for Visual Tracking}

\author{Wenzhang Zhou$^{1}$\thanks{Both authors contributed equally to this work.}, Longyin Wen$^{2\ast}$, Libo Zhang$^{3}$\thanks{Corresponding author (libo@iscas.ac.cn). This work is supported by the National Natural Science Foundation of China under Grant No. 61807033, the Key Research Program of Frontier Sciences, CAS, Grant No. ZDBS-LY-JSC038, Youth Innovation Promotion Association CAS, and Outstanding Youth Scientist Project of ISCAS.}\\
Dawei Du$^4$, Tiejian Luo$^1$, Yanjun Wu$^3$\\
$^1$University of Chinese Academy of Sciences, China, \\
$^2$JD Finance America Corporation, Mountain View, USA,\\
$^3$Institute of Software Chinese Academy of Sciences, China,\\
$^4$University at Albany, State University of New York, USA.\\
{\tt\small zhouwenzhang19@mails.ucas.ac.cn, longyin.wen.cv@gmail.com}\\
{\tt\small \{libo,yanjun\}@iscas.ac.cn, cvdaviddo@gmail.com}\\
}

\maketitle

\begin{abstract}
In this paper, we present a novel siamese motion-aware network (SiamMan) for visual tracking, which consists of the siamese feature extraction subnetwork, followed by the classification, regression, and localization branches in parallel. The classification branch is used to distinguish the foreground from background, and the regression branch is adopt to regress the bounding box of target. To reduce the impact of manually designed anchor boxes to adapt to different target motion patterns, we design the localization branch, which aims to coarsely localize the target to help the regression branch to generate accurate results. Meanwhile, we introduce the global context module into the localization branch to capture long-range dependency for more robustness in large displacement of target. In addition, we design a multi-scale learnable attention module to guide these three branches to exploit discriminative features for better performance. The whole network is trained offline in an end-to-end fashion with large-scale image pairs using the standard SGD algorithm with back-propagation. Extensive experiments on five  challenging benchmarks, \ie, VOT2016, VOT2018, OTB100, UAV123 and LTB35, demonstrate that SiamMan achieves leading accuracy with high efficiency. Code can be found at \url{https://isrc.iscas.ac.cn/gitlab/research/siamman}.
\end{abstract}

\section{Introduction}
Visual object tracking is a hot research direction with a wide range of applications, such as surveillance, autonomous driving, and human-computer interaction. Although significant progress has been made in recent years, it is still a challenging task due to various factors, including occlusion, abrupt motion, and illumination variation.

Modern visual tracking algorithms can be roughly divided into two categories: (1) the correlation filter based approaches \cite{DBLP:conf/cvpr/DanelljanBKF17,DBLP:conf/cvpr/LiTZ0018,DBLP:conf/cvpr/Sun0L018,DBLP:journals/tip/XuFWK19}, and (2) the deep convolution network based approaches \cite{DBLP:conf/cvpr/TaoGS16,DBLP:journals/corr/abs-1812-11703,DBLP:journals/corr/abs-1812-05050,DBLP:journals/corr/abs-1811-07628}. The correlation filter (CF) based approach trains a regressor for tracking using circular correlation via Fast Fourier Transform (FFT). With the arrival of the deep convolution network, some researchers use offline learned deep features \cite{DBLP:conf/eccv/DanelljanRKF16,DBLP:conf/cvpr/DanelljanBKF17,DBLP:conf/eccv/HeldTS16} to improve the accuracy. Considering efficiency, those trackers abandon model update in tracking process, which greatly harms the accuracy and generally performs worse than the CF based approaches.

Recently, the deep Siamese-RPN method \cite{DBLP:conf/cvpr/LiYWZH18} is presented, which formulates the tracking task as the one-shot detection task, \ie, using the bounding box in the first frame as the only exemplar. By exploiting the domain specific information, Siamese-RPN surpasses the performance of the CF based methods. Some methods \cite{DBLP:journals/corr/abs-1812-11703,DBLP:journals/corr/abs-1907-03892,DBLP:journals/corr/abs-1812-05050} attempt to improve the method \cite{DBLP:conf/cvpr/LiYWZH18} by using layer-wise and depth-wise feature aggregations, simultaneously predicting target bounding box and class-agnostic binary segmentation, and using ellipse fitting to estimate the bounding box rotation angle and size for better performance. However, the aforementioned methods rely on the pre-set anchor boxes to regress the bounding box of target, which can not adapt to various motion patterns and scales of targets, especially when the {\em fast motion} and {\em occlusion} challenges occur.

To that end, in this paper, we present a siamese motion-aware network (SiamMan) for visual tracking, which is formed by the siamese feature extraction subnetwork, followed by three paralleling branches, \ie, the classification, regression, and localization branches. Similar to \cite{DBLP:conf/cvpr/LiYWZH18}, the classification branch is used to distinguish the foreground from background, while the regression branch is used to regress the bounding box of target. To reduce the impact of manually designed anchor boxes to adapt to different motion patterns and scales of targets, we design a localization branch, which coarsely localizes the target to help the regression branch to generate more accurate results. Meanwhile, we introduce the global context module \cite{DBLP:journals/corr/abs-1904-11492} into the localization branch to capture long-range dependency, which makes the tracker to be more robust to the large target displacement. In addition, we also design a multi-scale learnable attention module to guide these three branches to exploit discriminative features for better performance. The whole network is trained in an end-to-end fashion offline with the large-scale image pairs by the standard SGD algorithm with back-propagation \cite{DBLP:journals/neco/LeCunBDHHHJ89} in the training sets of MS COCO \cite{DBLP:conf/eccv/LinMBHPRDZ14}, ImageNet DET/VID \cite{DBLP:journals/ijcv/RussakovskyDSKS15}, and YouTube-BoundingBoxes \cite{DBLP:conf/cvpr/RealSMPV17} datasets. For inference, the visual object tracking is formulated as the local one-shot detection task by using the bounding box of target in the first frame as the only exemplar. Several experiments are conducted on five challenging benchmarks, \ie, VOT2016 \cite{DBLP:conf/eccv/KristanLMFPCVHL16}, VOT2018 \cite{DBLP:conf/eccv/KristanLMFPZVBL18}, OTB2015 \cite{DBLP:journals/pami/WuLY15}, UAV123 \cite{DBLP:conf/eccv/MuellerSG16} and LTB35 \cite{DBLP:journals/corr/abs-1804-07056}. Our SiamMan method sets a new state-of-the-art on four datasets, \ie, VOT2016, VOT2018, OTB2015, and LTB35, and performs on par with the state-of-the-art on UAV123. Notably, it achieves $0.513$ and $0.462$ EAOs, improving $0.042$ and $0.016$ absolute values, \ie, $8.9\%$ and $3.6\%$ relative improvements, compared to the second best tracker on VOT2016 and VOT2018. Moreover,  ablation experiments are conducted to verify the effectiveness of different components in our method.

The main contributions of this work are summarized as follows. 
\begin{itemize*}
\item We propose a new siamese motion-aware network (SiamMan) for visual tracking, which is formed by a backbone feature extractor and three branches, \ie, the classification, regression, and localization branches. 
\item To capture the long-range dependency, we integrate the global context module \cite{DBLP:journals/corr/abs-1904-11492} into the localization branch, making the tracker to be more robust to large target displacement. 
\item We design a multi-scale learnable attention module to guide the network to exploit discriminative features for accurate results. 
\item SiamMan achieves the state-of-the-art results on four challenging benchmarks, \ie, VOT2016, VOT2018, OTB2015, and LTB35, and performs on par with the state-of-the-art on UAV123.
\end{itemize*}

\begin{figure*}[t]
\centering
\includegraphics[width=1.0\linewidth]{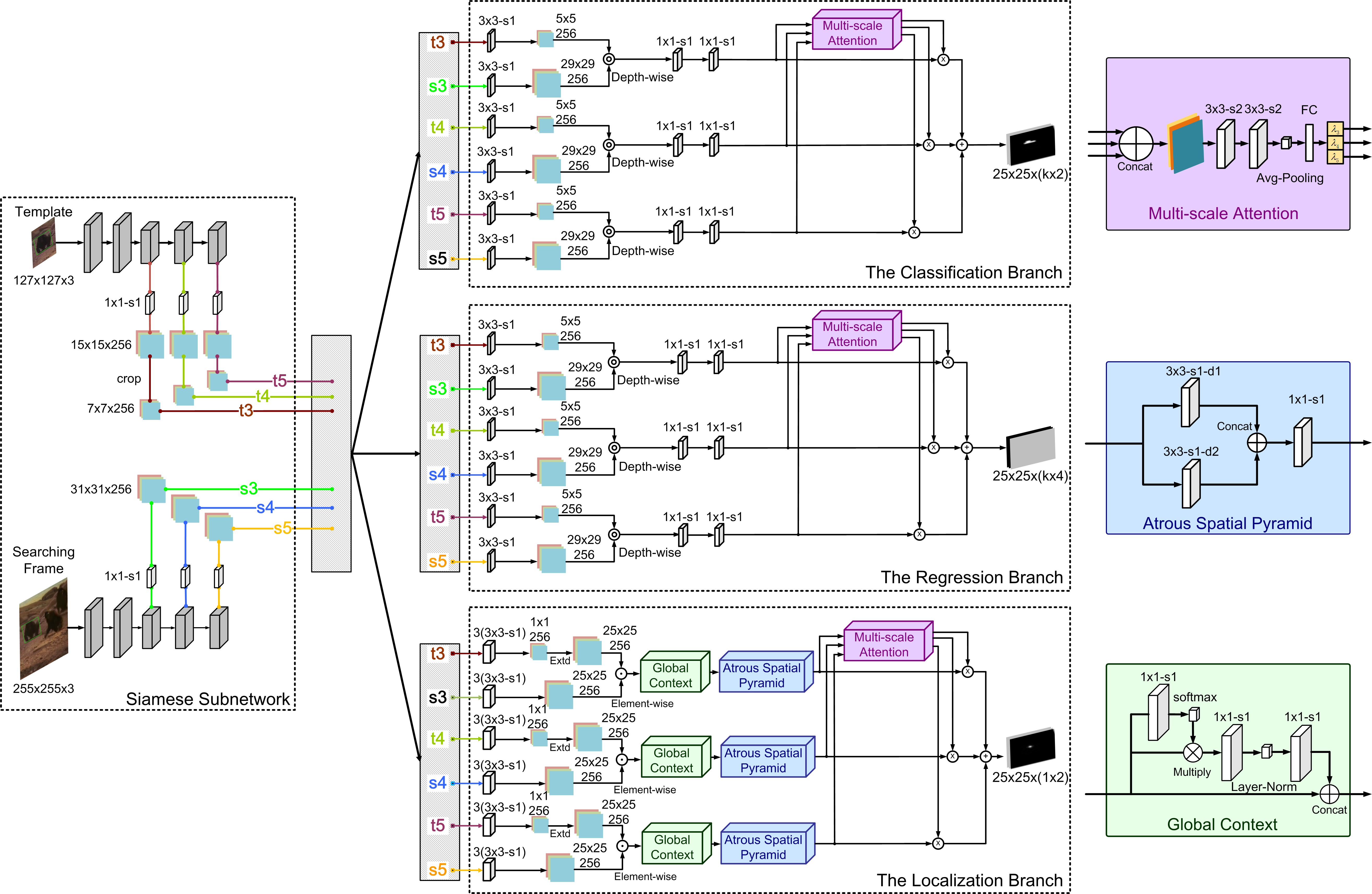}
\caption{The architecture of our SiamMan method, which consists of the siamese feature extraction subnetwork followed by the classification, regression, and localization branches in parallel. The pairs of feature maps from different layers of the siamese feature extraction subnetwork are fed into the three branches. ``3x3-s1-d2'' denotes a convolution layer with $3\times3$ kernel, stride $1$ and dilation rate $2$. Best view in color.}
\label{fig:architecture}
\end{figure*}

\section{Related work}
Visual tracking aims to estimate the states, \ie, sizes and locations, of target in the video sequence, with the given state in the first frame, which is an important and fundamental problems in computer vision community. Correlation filter based approach attracts much attention of researchers due to its computational efficiency and competitive performance \cite{DBLP:conf/eccv/DanelljanRKF16,DBLP:conf/cvpr/DanelljanBKF17,DBLP:conf/cvpr/LiTZ0018}. In recent years, the focus of researchers shift to the deep neural network based methods, such as MDNet \cite{DBLP:conf/cvpr/NamH16}, ATOM \cite{DBLP:journals/corr/abs-1811-07628}, SINT \cite{DBLP:conf/cvpr/TaoGS16}, SiamFC \cite{DBLP:conf/eccv/BertinettoVHVT16}, and SiamRPN \cite{DBLP:conf/cvpr/LiYWZH18}. MDNet \cite{DBLP:conf/cvpr/NamH16} learn the share representation of targets from multiple annotated video sequences for tracking, which has separate branches of domain-specific layers for binary classification at the end of the network, and shares the common information captured from all sequences in the preceding layers for generic representation learning. ATOM \cite{DBLP:journals/corr/abs-1811-07628} is formed by two components, \ie, a target estimation module, and a target classification module. The target estimation module is trained offline to predict the intersection over union overlap between the target and an estimated bounding box, and the target classification module is learned online to provide high robustness against distractor objects in the scene.

Some other researchers attempt to use the Siamese network for visual tracking. SINT \cite{DBLP:conf/cvpr/TaoGS16} and SiamFC \cite{DBLP:conf/eccv/BertinettoVHVT16} formulate the visual tracking problem as the pairwise similarity learning of the target in consecutive frames using the Siamese network. Dong \etal \cite{DBLP:conf/eccv/DongS18} use the triplet loss to train the Siamese network to exploit discriminative features instead of the pairwise loss, which can mine the potential relationship among exemplar, \ie, positive and negative instances, and contains more elements for training. To take fully use of semantic information, He \etal \cite{DBLP:conf/cvpr/HeLTZ18} construct a twofold Siamese networks, which is composed of a semantic branch and an appearance branch, and each of them is a similarity-learning Siamese network. Abdelpakey and Shehat \cite{DBLP:journals/corr/abs-1908-07905} use semantic and objectness information and produce a class-agnostic using a ridge regression network for object tracking.

After that, inspired by Region Proposal Network (RPN) in object detection, SiamRPN \cite{DBLP:conf/cvpr/LiYWZH18} formulates visual tracking as a local one-shot detection task in inference, which uses the Siamese network for feature extraction and RPN for target classification and regression. Fan and Ling \cite{DBLP:journals/corr/abs-1812-06148} construct a cascaded RPN from deep high-level to shallow low-level layers in a Siamese network. Zhu \etal \cite{DBLP:conf/eccv/ZhuWLWYH18} design a distractor-aware Siamese networks for accurate long-term tracking by using an effective sampling strategy to control the distribution of training data, and make the model focus on the semantic distractors. SiamRPN++ \cite{DBLP:journals/corr/abs-1812-11703} is improved from SiamRPN \cite{DBLP:conf/cvpr/LiYWZH18} by performing layer-wise and depth-wise aggregations, which not only improves the accuracy but also reduces the model size. Zhang and Peng \cite{DBLP:journals/corr/abs-1901-01660} design a residual network for visual tracking with controlled receptive field size and network stride. Han \etal \cite{DBLP:journals/access/HanDLSL19} introduce the anchor-free detection network into visual tracking directly. Moreover, SiamMask \cite{DBLP:journals/corr/abs-1812-05050} combines the fully-convolutional Siamese tracker with a binary segmentation head for accurate tracking. To track the rotated target accurately, Chen \etal \cite{DBLP:journals/corr/abs-1907-03892} improve SiamMask \cite{DBLP:journals/corr/abs-1812-05050} by using ellipse fitting to estimate the bounding box rotation angle and size with the mask on the target. However, the aforementioned algorithms fail to consider the variations of target motion patterns, resulting in failures when the {\em fast motion}, {\em occlusion}, and {\em camera motion} challenges occur.

\section{Siamese Motion-aware Network}
As shown in Figure \ref{fig:architecture}, our siamese motion-aware network is a feed-forward network, which is formed by a siamese feature extraction subnetwork, followed by three paralleling branches, \ie, the classification branch, the regression branch, and the localization branch. The classification branch is designed to distinguish the foreground proposals from the background, and the regression branch is used to regress the bounding box of target based on the preset anchor boxes. Inspired by \cite{DBLP:journals/corr/abs-1904-07850}, we integrate a localization branch, used to coarsely localize the target to help the regression branch to adapt to different motion patterns. Let $k$ be the number of pre-set anchors. Thus, we have $2k$ channels for classification, $4k$ channels for regression and $2$ channels for localization, and denote the output feature maps of the three branches as ${\cal O}_{w\times{h}\times{2k}}^{\text{cls}}$, ${\cal O}_{w\times{h}\times{4k}}^{\text{reg}}$, and ${\cal O}_{w\times{h}\times{2}}^{\text{loc}}$. Notably, each point in ${\cal O}_{w\times{h}\times{2k}}^{\text{cls}}$, ${\cal O}_{w\times{h}\times{4k}}^{\text{reg}}$, and ${\cal O}_{w\times{h}\times{2}}^{\text{loc}}$ contains $2k$, $4k$, and $2$ channel vectors, representing the positive and negative activations of each anchor at the corresponding locations of original map for each branch. In the following sections, we will describe each module in our SiamMan in detail.

{\flushleft {\bf Siamese feature extraction subnetwork}.}
Inspired by \cite{DBLP:conf/cvpr/LiYWZH18}, we use the fully convolution network without padding is used in the Siamese feature extraction subnetwork. Specifically, there are two components in the Siamese feature extraction subnetwork, \ie, the {\em template module} encoding the target patch in the historical frame, and the {\em detection module} encoding the image patch including the target in the current frame. The two components share parameters in CNN. Let $\alpha$ be the target patch fed into the template module, and $\beta$ be the image patch fed into the detection module. We denote the output feature maps of the Siamese feature extraction subnetwork at the $i$-th layer as $\phi_{i}(\alpha)$ and $\phi_{i}(\beta)$. Then, we split each of them into three branches, \ie, $\phi^{\text{cls}}_{i}(\alpha)$ and $\phi^{\text{cls}}_{i}(\beta)$ for the classification branch, $\phi^{\text{reg}}_{i}(\alpha)$ and $\phi^{\text{reg}}_{i}(\beta)$ for the regression branch, and $\phi^{\text{loc}}_{i}(\alpha)$ and $\phi^{\text{loc}}_{i}(\beta)$ for the localization branch, by a convolution layer with kernel size $3\times3$ and stride $1$, but keeping the number of channels unchanged. Similar to the previous work \cite{DBLP:journals/corr/abs-1812-11703}, we use the ResNet-50 network \cite{DBLP:conf/cvpr/HeZRS16} as the backbone. To reduce the computational complexity, we extract the feature maps from the backbone with the channel $256$ by one $1\times1$ convolutional layer, and then crop the center $7\times7$ regions \cite{DBLP:conf/cvpr/ValmadreBHVT17} from the $15\times15$ feature maps as the template feature. Due to the paddings of all layers in the backbone, the $7\times7$ feature map can still represent the entire target region.

{\flushleft {\bf Classification branch}.}
As shown in Figure \ref{fig:architecture}, the classification branch takes the multi-scale features produced by the {\em template} and {\em detection} modules of the Siamese feature extraction subnetwork, \eg, $\text{t3}$, $\text{s3}$, $\text{t4}$, $\text{s4}$, $\text{t5}$, and $\text{s5}$, to compute the correlation feature maps between the input {\em template} ($\phi^{\text{cls}}_{i}(\alpha)$) and {\em detection} ($\phi^{\text{cls}}_{i}(\beta)$) feature maps, \ie,
\begin{equation}
\begin{array}{ll}
{\it F}_{w\times{h}\times{2k}}^{\text{cls}}(m)=\phi^{\text{cls}}_{m}(\alpha)\star\phi^{\text{cls}}_{m}(\beta),
\end{array}
\end{equation}
where $\star$ denotes depth-wise convolution operation. We use two convolution layers with the kernel size $1\times1$ and stride size $1$, to produce the features with $2k$ channels, \ie, ${\cal F}_{w\times{h}\times{2k}}^{\text{cls}}(m)$, $m=1,\cdots,L$, where $L$ is the total number of feature maps for prediction. After that, we use the multi-scale attention module to guide the branch to exploit discriminative features fore accurate results. Specifically, we first concatenate the feature maps at different layers, \ie, ${\cal F}_{w\times{h}\times{2k}}^{\text{cls}}(m)$, $m=1,\cdots,L$, and use two convolutional layers with the kernel size $3\times3$ and stride size $2$, followed by an average pooling and fully connected layers to produce the weights, \ie, $\gamma^{\text{cls}}_{i}$. After that, the feature maps with different scales are summed with the weights $\gamma^{\text{cls}}_{m}$ to generate the final predictions ${\cal O}_{w\times{h}\times{2k}}^{\text{cls}}$, \ie,
\begin{equation}
\begin{array}{ll}
{\cal O}_{w\times{h}\times{2k}}^{\text{cls}} &= \sum_{m=1}^{L}\gamma^{\text{cls}}_{m}\cdot{\cal F}_{w\times{h}\times{2k}}^{\text{cls}}(m).
\end{array}
\end{equation}
Each point in the output ${\cal O}_{w\times{h}\times{2k}}^{\text{cls}}$ is a $2k$ channel vector, indicating the positive and negative activations of each anchor at the corresponding locations of original map. Notably, the weights $\gamma^{\text{cls}}_{m}$, $m=1,\cdots,L$, are learned in the training phase, \ie, the gradients of the whole network can be back-propagated to update $\gamma^{\text{cls}}_{m}$, $m=1,\cdots,L$. Please see Figure \ref{fig:architecture} for more details.

{\flushleft {\bf Regression branch}.}
As described above, the regression branch is designed to generate the accurate bounding box of target in the current video frame. As shown in Figure  \ref{fig:architecture}, we compute the correlation feature maps between the input {\em template} and {\em detection} feature maps. For example, for the feature map at the $m$-th layer, the correlation feature map ${\it F}_{w\times{h}\times{4k}}^{\text{reg}}(m)$ is computed as
\begin{equation}
\begin{array}{ll}
{\it F}_{w\times{h}\times{4k}}^{\text{reg}}(m)=\phi^{\text{reg}}_{m}(\alpha)\star\phi^{\text{reg}}_{m}(\beta),
\end{array}
\end{equation}
where $\phi^{\text{reg}}_{m}(\alpha)$ and $\phi^{\text{reg}}_{m}(\beta)$ are the $m$-th feature map from the {\em template} and {\em detection} modules. After that, two convolution layers with the kernel size $1\times1$ and stride size $1$, are applied on ${\it F}_{w\times{h}\times{4k}}^{\text{reg}}(m)$ to produce the corresponding feature map ${\cal F}_{w\times{h}\times{4k}}^{\text{reg}}(m)$, $m=1,\cdots,L$, keeping the channel size $4k$ unchanged, where $L$ is the total number of feature maps used for prediction. Similar to the {\em classification branch}, we use the multi-scale attention module with the learnable weights $\gamma^{\text{reg}}_{m}$, $m=1,\cdots,L$, to make the branch focus on exploiting discriminative features to generate accurate results, \ie,
\begin{equation}
\begin{array}{ll}
{\cal O}_{w\times{h}\times{4k}}^{\text{reg}} &= \sum_{m=1}^{L}\gamma^{\text{reg}}_{m}\cdot{\cal F}_{w\times{h}\times{4k}}^{\text{reg}}(m),
\end{array}
\end{equation}
where ${\cal O}_{w\times{h}\times{4k}}^{\text{reg}}$ is the output of the regression branch. Each point on ${\cal O}_{w\times{h}\times{4k}}^{\text{reg}}$ contains a $4k$ channel vector, indicating the normalized distance between the predicted anchor box and the ground-truth bounding box.

{\flushleft {\bf Localization branch}.}
In the visual tracking task, different targets have different motion patterns, \ie, some targets move fast, while some targets move slowly. The {\em regression branch} relies on pre-set anchor boxes are inaccurate in challenging scenarios, such as {\em fast motion} and {\em small object}. To make our tracker adapt to various scales and motion patterns of targets, we introduce a {\em localization branch}, which is used to coarsely localizes the target to help the {\em regression branch} to produce accurate results. Specifically, taking the multi-scale features $\phi^{\text{loc}}_{i}(\alpha)$ and $\phi^{\text{loc}}_{i}(\beta)$ from the Siamese feature extraction subnetwork, we compute the correlation feature map as
\begin{equation}
\begin{array}{ll}
{\it F}_{w\times{h}\times2}^{\text{loc}}(m)=\mathbb{E}[\phi^{\text{loc}}_{m}(\alpha)]\odot\phi^{\text{loc}}_{m}(\beta),
\end{array}
\end{equation}
where $\mathbb{E}[\cdot]$ denotes the resize operation to make the two feature maps to be the same size, and $\odot$ denotes element-wise multiplication operation, see Figure \ref{fig:architecture}. After that, we insert the global context module \cite{DBLP:journals/corr/abs-1904-11492} to integrate long-range dependency between target and background regions, making the tracker to be more robust to the large target displacement. Inspired by \cite{DBLP:journals/pami/ChenPKMY18}, we design the atrous spatial pyramid module to capture the context information at multiple scales, which applies two parallel atrous convolution with different rates, followed by a convolution layer with $1\times1$ kernel size and stride $1$. In this way, we can generate the multi-scale discriminative features ${\cal F}_{w\times{h}\times{2}}^{\text{loc}}(m)$, where $m=1,\cdots,L$. Then, similar to the {\em classification} and {\em regression} branches, we use the multi-scale attention module with the learnable weights $\gamma^{\text{loc}}_{m}$, $m=1,\cdots,L$ to generate the final predictions. That is,
\begin{equation}
\begin{array}{ll}
{\cal O}_{w\times{h}\times{2}}^{\text{loc}} &= \sum_{m=1}^{L}\gamma^{\text{loc}}_{m}\cdot{\cal F}_{w\times{h}\times{2}}^{\text{loc}}(m).
\end{array}
\end{equation}
Notably, each point on the prediction ${\cal O}_{w\times{h}\times{2}}^{\text{loc}}$ is a two channel vector, representing the offset of the corresponding center location in original map.

{\flushleft {\bf Loss function}.}
The loss function in our method is formed by three terms corresponding to three branches, \ie, the classification loss ${\it L}_{\text{cls}}$, the regression loss ${\it L}_{\text{reg}}$, and the localization loss ${\it L}_{\text{loc}}$. The overall loss function is computed as:
\begin{equation}
{\cal L} = \lambda_\text{cls}{\it L}_\text{cls}(\mathrm{u}, \mathrm{u}^\ast)+\lambda_\text{reg}{\it L}_\text{reg}(\mathrm{p}, \mathrm{p}^\ast)+\lambda_\text{loc}{\it L}_\text{loc}(\mathrm{c}, \mathrm{c}^\ast),
\label{equ:loss}
\end{equation}
where $\lambda_\text{cls}$, $\lambda_\text{reg}$ and $\lambda_\text{loc}$ are the parameters used to balance the three loss terms. $\mathrm{u}$ and $\mathrm{u}^\ast$ are the predicted and ground-truth labels of the target bounding boxes, $\mathrm{p}$ and $\mathrm{p}^\ast$ are the predicted and ground-truth bounding boxes, and $\mathrm{c}$ and $\mathrm{c}^\ast$ are the predicted and ground-truth labels of the center of target. We use the cross-entropy loss to supervise the {\em classification} and {\em localization} branches, and L1 loss to supervise for the {\em regression} branch.

Specifically, the classification loss ${\it L}_{\text{cls}}(\mathrm{u}, \mathrm{u}^\ast)$ is computed as
\begin{equation}
\begin{array}{ll}
{\it L}_{\text{cls}}(\mathrm{u}, u^\ast)&=-\frac{1}{2}\sum_{i}\sum_{j}\sum_{k}\big(u_{i,j,k}^\ast\log{u_{i,j,k}} \\
&+ (1-u_{i,j,k}^\ast)\log(1-u_{i,j,k}) \big),
\end{array}
\label{equ:cla_loss}
\end{equation}
where $u_{i,j,k}^\ast$ is the ground-truth label of the $k$-th anchor at $(i,j)$ of the output ${\cal O}_{w\times{h}\times{2k}}^{\text{cls}}$, and $u_{i,j,k}$ is the predicted label of the $k$-th anchor at $(i,j)$ generated by the softmax operation from ${\cal O}_{w\times{h}\times{2k}}^{\text{cls}}$ over $2$ categories.

Meanwhile, we use the L1 loss function to compute the regression loss ${\it L}_{\text{reg}}(p, p^\ast)$, \ie,
\begin{equation}
\begin{array}{ll}
{\it L}_{\text{reg}}(\mathrm{p}, \mathrm{p}^\ast)=\frac{1}{N}\sum_i\sum_j\sum_k[u_{i,j,k}^\ast > 0]\|\delta(p_{i,j,k},p_{i,j,k}^\ast)\|_1,
\end{array}
\end{equation}
where $N$ is the number of positive anchors, and the Iverson bracket indicator function $[u_{i,j,k}^\ast > 0]$ outputs $1$ when the condition is true, \ie, the anchor is not negative $u_{i,j,k}^\ast > 0$, and $0$ otherwise. $p_{i,j,k}=(x, y, w, h)$ and $p_{i,j,k}^\ast=(x^\ast, y^\ast, w^\ast, h^\ast)$ are the predicted and ground-truth bounding boxes, where $(x, y)$ and $(x^\ast, y^\ast)$ are the center coordinates and $(w, h)$ and $(w^\ast, h^\ast)$ are the sizes. We use the normalized distances $\delta$ to compute the regression loss, \ie, $\delta(p,p^\ast)=((x^\ast-x)/x,(y^\ast-y)/y,\ln(w^\ast/w),\ln(h^\ast/h))$.

Moreover, we also use the cross-entropy loss for the localization branch as follows.
\begin{equation}
\begin{array}{ll}
{\it L}_{\text{loc}}(\mathrm{c}, \mathrm{c}^\ast)&=-\frac{1}{2}\sum_{i}\sum_{j}\big(c_{i,j}^\ast\log{c_{i,j}}\\
&+(1-c_{i,j}^\ast)\log{(1-c_{i,j})}\big),
\end{array}
\end{equation}
where $c_{i,j}^\ast$ is the ground-truth label of the center of target at $(i,j)$ of the output ${\cal O}_{w\times{h}\times{2}}^{\text{loc}}$, and $c_{i,j}$ is the predicted label of the center of target at $(i,j)$ generated by the softmax operation from ${\cal O}_{w\times{h}\times{2}}^{\text{loc}}$. Notably, we generate the ground-truth center location of the target $\mathrm{c}^\ast$ (where $c_{i,j}^\ast\in[0,1]$) using the Gaussian kernel with the object size-adaptive standard deviation \cite{DBLP:conf/eccv/LawD18}.

\subsection{Training and Inference}
{\flushleft {\bf Data augmentation}.} We use several data augmentation strategies such as blur, rescale, rotation, flipping and gray scaling to construct a robust model to adapt to variations of objects using the video sequences in MS COCO \cite{DBLP:conf/eccv/LinMBHPRDZ14}, ImageNet DET/VID \cite{DBLP:journals/ijcv/RussakovskyDSKS15}, and YouTube-BoundingBoxes \cite{DBLP:conf/cvpr/RealSMPV17}. For the positive image pairs, we randomly select two frames from the same video sequences with the interval less than $100$ frames, or different image patches including target object in the MS COCO and ImageNet DET datasets. Meanwhile, for the negative image pairs, we randomly select an image from the datasets and another one without including the same target. Notably, the ratio between the positive and negative pairs is set to $4:1$.

{\flushleft {\bf Anchors design and matching}.}
For each point, we pave $5$ anchors with stride $8$ on each pixel, where the anchor ratios are set to $[1/3, 1/2, 1/1, 2/1, 3/1]$ and the anchor scale is set to $8$. Meanwhile, during the training phase, we determine the correspondence between the anchors and ground-truth boxes based on the jaccard overlap. Specifically, if the overlap between the anchor and ground-truth box is larger than $0.6$, the anchor is determined as positive. Meanwhile, if the overlap between the anchor and all ground-truth boxes is less than $0.3$, the anchor is determined as negative.

{\flushleft {\bf Optimization}.}
The whole network is trained in an end-to-end manner using the SGD optimization algorithm with $0.9$ momentum and $0.0001$ weight decay on the training sets of MS COCO \cite{DBLP:conf/eccv/LinMBHPRDZ14}, ImageNet DET/VID \cite{DBLP:journals/ijcv/RussakovskyDSKS15}, and YouTube-BoundingBoxes \cite{DBLP:conf/cvpr/RealSMPV17} datasets. Notably, we use three stages to train the proposed method empirically. For the first two stages in the training process, we disable the multi-scale attention modules in the three branches, \ie, set equal weights to different scales of features.
\begin{itemize*}
\item In the first stage, the backbone ResNet-50 network in the siamese feature subnetwork is initialized by the pre-trained model on the ILSVRC CLS-LOC dataset \cite{DBLP:journals/ijcv/RussakovskyDSKS15}. We train the classification and regression branches in the first $10$ epochs with other parameters fixed, and then train the siamese feature subnetwork, and the classification and regression branches in the next $10$ epochs.
\item In the second stage, we finetune the classification, regression and localization branches with other parameters fixed in the first $10$ epochs, and then train the whole network in the next $10$ epochs.
\item In the third stage, we enable the multi-attention module and learn the weights of different scales of features with other parameters fixed in the first $15$ epochs. After that, the whole network is finetuned in the next $5$ epochs.
\end{itemize*}
In each stage, we set the initial learning rate to $0.001$, and gradually increase it to $0.005$ in the first $5$ epochs. We decrease it to $0.0005$ in the next $15$ epochs.

{\flushleft {\bf Inference}.}
In the inference phase, our tracker takes the current video frame and the template target patch as input, and outputs the classification, regression, and localization results. Then, we perform softmax operation on both the outputs of the classification and localization results to obtain the positive activations, \ie, $\mathrm{u}$ with the size $w\times{h}\times{k}$, and $\mathrm{c}$ with the size $w\times{h}\times{1}$. After that, we expand the localization result $\mathrm{c}$ to make it to the same size of the classification result $\mathrm{u}$. In this way, the final prediction is computed by the weight combination of four terms, \ie, the localization result $\mathrm{c}$, the classification result $\mathrm{u}$, the cosine window $\xi$ with the size $w\times{h}$ (expanding to $w\times{h}\times{k}$), and the scale change penalty $\rho$ with the size $w\times{h}\times{k}$ \cite{DBLP:conf/cvpr/LiYWZH18},
\begin{equation}
\begin{aligned}
\Theta_{w\times{h}\times{k}} = \omega_2\cdot\rho\cdot\big(\omega_1\cdot\mathrm{u} + (1-\omega_1)\cdot\mathrm{c}\big)
+ (1-\omega_2)\cdot\xi,
\end{aligned}
\label{equ:center}
\end{equation}
Notably,
The cosine window $\xi$ is used to suppress the boundary outliers \cite{DBLP:journals/pami/HenriquesC0B15}, and the scale change penalty $\rho$ to suppress large change in size and ratio \cite{DBLP:conf/cvpr/LiYWZH18}. The weights $\omega_1$ and $\omega_2$ are used to balance the above terms, which are set to $0.7$ and $0.6$, empirically. After that, we can obtain the optimal center location and scale of target based on the maximal score on $\Theta_{w\times{h}\times{k}}$. Notably, the target size is updated by the linear interpolation to guarantee the smoothness of size.

\section{Experiment}
Our SiamMan method is implemented using the Pytorch tracking platform PySOT\footnote{\url{https://github.com/STVIR/pysot}}. Several experiments are conducted on five challenging datasets, \ie, VOT2016 \cite{DBLP:conf/eccv/KristanLMFPCVHL16}, VOT2018 \cite{DBLP:conf/eccv/KristanLMFPZVBL18}, OTB100 \cite{DBLP:journals/pami/WuLY15}, UAV123 \cite{DBLP:conf/eccv/MuellerSG16} and LTB35 \cite{DBLP:journals/corr/abs-1804-07056}, to demonstrate the effectiveness of the proposed method. All experiments are conducted on a workstation with the Intel i7-7800X CPU, 8G memory, and $2$ NVIDIA RTX2080 GPUs. The average tracking speed is $45$ fps. The source code and models will be released after the paper is accepted.

{\flushleft {\bf Evaluation protocol}.}
For the VOT2016 \cite{DBLP:conf/eccv/KristanLMFPCVHL16} and VOT2018 \cite{DBLP:conf/eccv/KristanLMFPZVBL18} datasets, we use the evaluation protocol in the VOT Challenge \cite{DBLP:conf/eccv/KristanLMFPCVHL16,DBLP:conf/eccv/KristanLMFPZVBL18}, \ie, the {\em Expected Average Overlap} (EAO), {\em Accuracy} (A), and {\em Robustness} (R) are used to evaluate the performance of trackers. The {\em Accuracy} score indicates the average overlap of the successfully tracked frames, and the {\em Robustness} score indicates the failure rate of the tracking frames\footnote{We define the failure of tracking if the overlap between the tracking result and ground-truth is reduced to $0$.}. EAO takes both accuracy and robustness into account, which is used as the primary metric for ranking trackers.

Meanwhile, for the OTB100 \cite{DBLP:journals/pami/WuLY15} and UAV123 \cite{DBLP:conf/eccv/MuellerSG16} datasets, we use the success and precision scores to evaluate the performance of trackers based on the evaluation methodology in \cite{DBLP:journals/pami/WuLY15}. The success score is defined as the area under the success plot, \ie, the percentage of successfully tracked frames\footnote{If the overlap between the predicted bounding box and ground-truth box in a frame is larger than a threshold, we regard the frame as a successfully tracked frames.} {\em vs.} bounding box overlap threshold in the interval $[0, 1]$. The precision score is computed as the percentage of frames whose predicted location is within a given distance threshold from the center of ground-truth box based on the Euclidean distance on the image plane. We set the distance threshold to $20$ pixels in our evaluation. In general, the success score is used as the primary metric for ranking trackers.

For the long-term tracking LTB35 dataset \cite{DBLP:journals/corr/abs-1804-07056}, we use three metrics including tracking precision ($\mathrm{P}$), tracking recall ($\mathrm{R}$) and tracking $\mathrm{F}$-score in evaluation. The tracking methods are ranked by the maximum $\mathrm{F}$-score over different confidence thresholds, \ie, $\mathrm{F}=\frac{2\mathrm{P}\cdot\mathrm{R}}{\mathrm{P}+\mathrm{R}}$.

\subsection{State-of-the-art Comparison}
We compare the proposed method to the state-of-the-art trackers on five challenging datasets. For a fair comparison, the tracking results of other trackers are directly taken from the published papers.

\begin{table}[t]
  \centering
  \caption{Comparisons with the state-of-the-art on VOT2016 \cite{DBLP:conf/eccv/KristanLMFPCVHL16} in terms of EAO, robustness, and accuracy. $\ast$ denotes that the result is obtained using the PySOT platform.}
  \label{tab:res_vot2016}
  \small \setlength{\tabcolsep}{8.0pt}
    \begin{tabular}{c|ccc}
    \hline
    Tracker & Accuracy & Robustness & EAO \\
    \hline
    C-COT \cite{DBLP:conf/eccv/DanelljanRKF16} &0.539 &0.238 &0.331 \\
    SiamRPN \cite{DBLP:conf/cvpr/LiYWZH18} & 0.560 & 1.080 & 0.344 \\
    FCAF \cite{DBLP:journals/access/HanDLSL19} & 0.581 & 1.020 & 0.356 \\
    C-RPN \cite{DBLP:journals/corr/abs-1812-06148}& 0.594 & 0.950 & 0.363 \\
    SiamRPN+ \cite{DBLP:journals/corr/abs-1901-01660}& 0.580 & 0.240 & 0.370 \\
    ECO \cite{DBLP:conf/cvpr/DanelljanBKF17} &0.550 &0.200 &0.375 \\
    ASRCF \cite{DBLP:conf/cvpr/Dai0LSL19} &0.563 &0.187 &0.391 \\
    DaSiamRPN \cite{DBLP:conf/eccv/ZhuWLWYH18} & 0.610 & 0.220 & 0.411 \\
    SiamMask$^\ast$ \cite{DBLP:journals/corr/abs-1812-05050}& \color{blue}{\bf 0.643} & 0.219 & 0.455 \\
    SiamRPN++$^\ast$ \cite{DBLP:journals/corr/abs-1812-11703}& 0.642 & 0.196 & 0.464 \\
    SiamMask\_E$^\ast$ \cite{DBLP:journals/corr/abs-1907-03892} & \color{red}{\bf 0.677} & 0.224 & 0.466 \\
    PTS \cite{DBLP:journals/corr/abs-1904-03280} & 0.642 & \color{red}{\bf 0.144} &\color{blue}{\bf 0.471} \\
    \hline
	\hline
    SiamMan$^\ast$ & 0.636 & \color{blue}{\bf 0.149}  &\color{red}{\bf 0.513} \\
    \hline
    \end{tabular}
\end{table}

{\flushleft {\bf VOT2016}.}
We conduct experiments on the VOT2016 dataset \cite{DBLP:conf/eccv/KristanLMFPCVHL16} to evaluate the performance of our SiamMan method in Table \ref{tab:res_vot2016}. VOT2016 contains $60$ sequences. Each sequence is per-frame annotated with the following visual attributes: {\em occlusion}, {\em illumination change}, {\em motion change}, {\em size change}, {\em camera motion}, and {\em unassigned}. As shown in Table \ref{tab:res_vot2016}, our SiamMan method achieves the best EAO score $0.513$ and the second best robustness score $0.149$. Notably, our method sets a new state-of-the-art by improving $0.042$ absolute value, \ie, $8.9\%$ relative improvement, compared to the second best tracker PTS \cite{DBLP:journals/corr/abs-1904-03280}. However, our method produce a relative lower accuracy score compared to  SiamMask \cite{DBLP:journals/corr/abs-1812-05050} and SiamMask\_E \cite{DBLP:journals/corr/abs-1907-03892}. The SiamMask and SiamMask\_E methods estimate the rotated bounding box based on the mask generated by the segmentation head, resulting in relative more accurate bounding box, especially for the non-rigid targets. Compared to SiamRPN++ \cite{DBLP:journals/corr/abs-1812-11703}, our SiamMan method produces $0.049$ and $0.005$ higher EAO and robustness scores, indicating that the localization branch can significantly decrease the tracking failure.

\begin{table}[t]
  \centering
  \caption{Comparison results on VOT2018 \cite{DBLP:conf/eccv/KristanLMFPZVBL18}. $\ast$ denotes that the result is obtained using the PySOT platform.}
   \label{tab:res_vot2018}
   \small \setlength{\tabcolsep}{8.0pt}
    \begin{tabular}{c|ccc}
    \hline
    Tracker & Accuracy & Robustness & EAO \\
    \hline
    SiamFC \cite{DBLP:conf/eccv/BertinettoVHVT16} &0.503 &0.585 &0.188 \\
    DSiam \cite{DBLP:conf/iccv/Guo0ZHWW17} &0.215 &0.646 &0.196 \\
    SiamRPN \cite{DBLP:conf/cvpr/LiYWZH18} &0.490 &0.460 &0.244 \\
    ECO \cite{DBLP:conf/cvpr/DanelljanBKF17} &0.484 &0.276 &0.280 \\
    SA\_Siam\_R \cite{DBLP:conf/cvpr/HeLTZ18} &0.566 &0.258 &0.337 \\
    DeepSTRCF \cite{DBLP:conf/cvpr/LiTZ0018} & 0.523 & 0.215 & 0.345 \\
    DRT \cite{DBLP:conf/cvpr/Sun0L018}  & 0.519 & 0.201 & 0.356 \\
    RCO \cite{DBLP:conf/eccv/KristanLMFPZVBL18}  & 0.507 & 0.155 & 0.376 \\
    UPDT \cite{DBLP:conf/eccv/BhatJDKF18}  & 0.536 & 0.184 & 0.378 \\
    DaSiamRPN \cite{DBLP:conf/eccv/ZhuWLWYH18} & 0.586 & 0.276 & 0.383 \\
    MFT \cite{DBLP:conf/eccv/KristanLMFPZVBL18}  & 0.505 & \color{red}{\bf 0.140} & 0.385 \\
    LADCF \cite{DBLP:journals/tip/XuFWK19} & 0.503 & 0.159 & 0.389 \\
    DomainSiam \cite{DBLP:journals/corr/abs-1908-07905} & 0.593 & 0.221 & 0.396 \\
    PTS \cite{DBLP:journals/corr/abs-1904-03280}  & 0.612 & 0.220 & 0.397 \\
    ATOM \cite{DBLP:journals/corr/abs-1811-07628}  & 0.590 & 0.204 & 0.401 \\
    SiamRPN++$^\ast$ \cite{DBLP:journals/corr/abs-1812-11703}& 0.601 & 0.234 & 0.415 \\
    SiamMask$^\ast$ \cite{DBLP:journals/corr/abs-1812-05050}& \color{blue}{\bf 0.615} & 0.248 & 0.423 \\
    DiMP-50 \cite{DBLP:journals/corr/abs-1904-07220}  & 0.597 & \color{blue}{\bf 0.153} & 0.440 \\
    SiamMask\_E$^\ast$ \cite{DBLP:journals/corr/abs-1907-03892}& \color{red}{\bf 0.655} & 0.253 &\color{blue}{\bf 0.446} \\
    \hline
	\hline
    SiamMan$^\ast$ & 0.605 & 0.183 &\color{red}{\bf 0.462} \\
    \hline
    \end{tabular}
\end{table}

{\flushleft {\bf VOT2018}.}
The VOT2018 dataset consists of $60$ challenging video sequences, which is annotated with the same standard as VOT2016 \cite{DBLP:conf/eccv/KristanLMFPCVHL16}. We evaluate the proposed SiamMan method on VOT2018 \cite{DBLP:conf/eccv/KristanLMFPZVBL18}, and report the results in Table \ref{tab:res_vot2018}. As shown in Table \ref{tab:res_vot2018}, our SiamMan method outperforms the state-of-the-art methods in terms of the primary ranking metric EAO. SiamMask\_E \cite{DBLP:journals/corr/abs-1907-03892} and SiamMask \cite{DBLP:journals/corr/abs-1812-05050} estimate the rotated bounding boxes of targets based on the segmentation results, producing higher accuracy scores, \ie, $0.655$ and $0.615$. Our SiamMan method outperforms SiamRPN++ \cite{DBLP:journals/corr/abs-1812-11703}, \ie, improving $0.047$ ($0.462$ {\em vs.} $0.415$) EAO and $0.051$ ($0.183$ {\em vs.} $0.234$) robustness, which fully demonstrates the effectiveness of the designed localization branch and multi-scale attention module.

\begin{figure}[t]
\centering
\includegraphics[width=\linewidth]{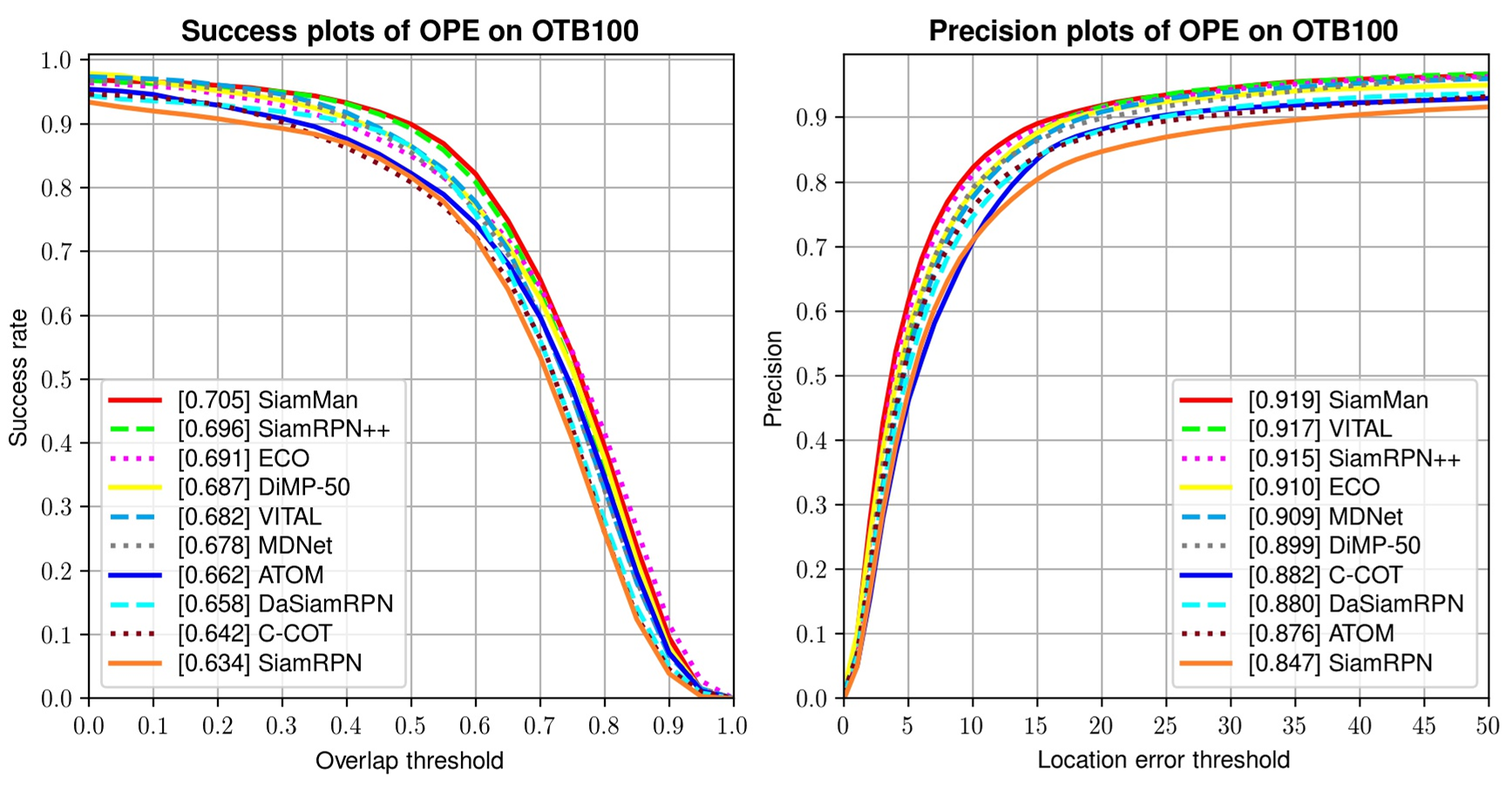}
\caption{Success and precision plots on the OTB100 dataset \cite{DBLP:journals/pami/WuLY15}.}
\label{fig:otb}
\end{figure}

{\flushleft {\bf OTB100}.}
OTB100 \cite{DBLP:journals/pami/WuLY15} is a challenging dataset, which consists of $100$ video sequences. We compare our SiamMan method with several representative trackers, \ie, SiamRPN++ \cite{DBLP:journals/corr/abs-1812-11703}, ECO \cite{DBLP:conf/cvpr/DanelljanBKF17}, DiMP-50 \cite{DBLP:journals/corr/abs-1904-07220}, VITAL \cite{DBLP:conf/cvpr/Song0WGBZSL018}, MDNet \cite{DBLP:conf/cvpr/NamH16}, ATOM \cite{DBLP:journals/corr/abs-1811-07628}, DaSiamRPN \cite{DBLP:conf/eccv/ZhuWLWYH18}, C-COT \cite{DBLP:conf/eccv/DanelljanRKF16}, and SiamRPN \cite{DBLP:conf/cvpr/LiYWZH18}, shown in Figure \ref{fig:otb}. As shown in Figure \ref{fig:otb}, our method achieves the best performance in both success and precision scores, \ie, $0.705$ success score and $0.919$ precision score. VITAL \cite{DBLP:conf/cvpr/Song0WGBZSL018} achieves the second best precision score $0.917$ but much worse success score $0.682$. Compared to SiamRPN++ \cite{DBLP:journals/corr/abs-1812-11703}, our method improves $0.009$ in success score (\ie, $0.705$ {\em vs.} $0.696$) and $0.004$ in precision score ($0.919$ {\em vs.} $0.915$).

\begin{figure}[t]
\centering
\includegraphics[width=\linewidth]{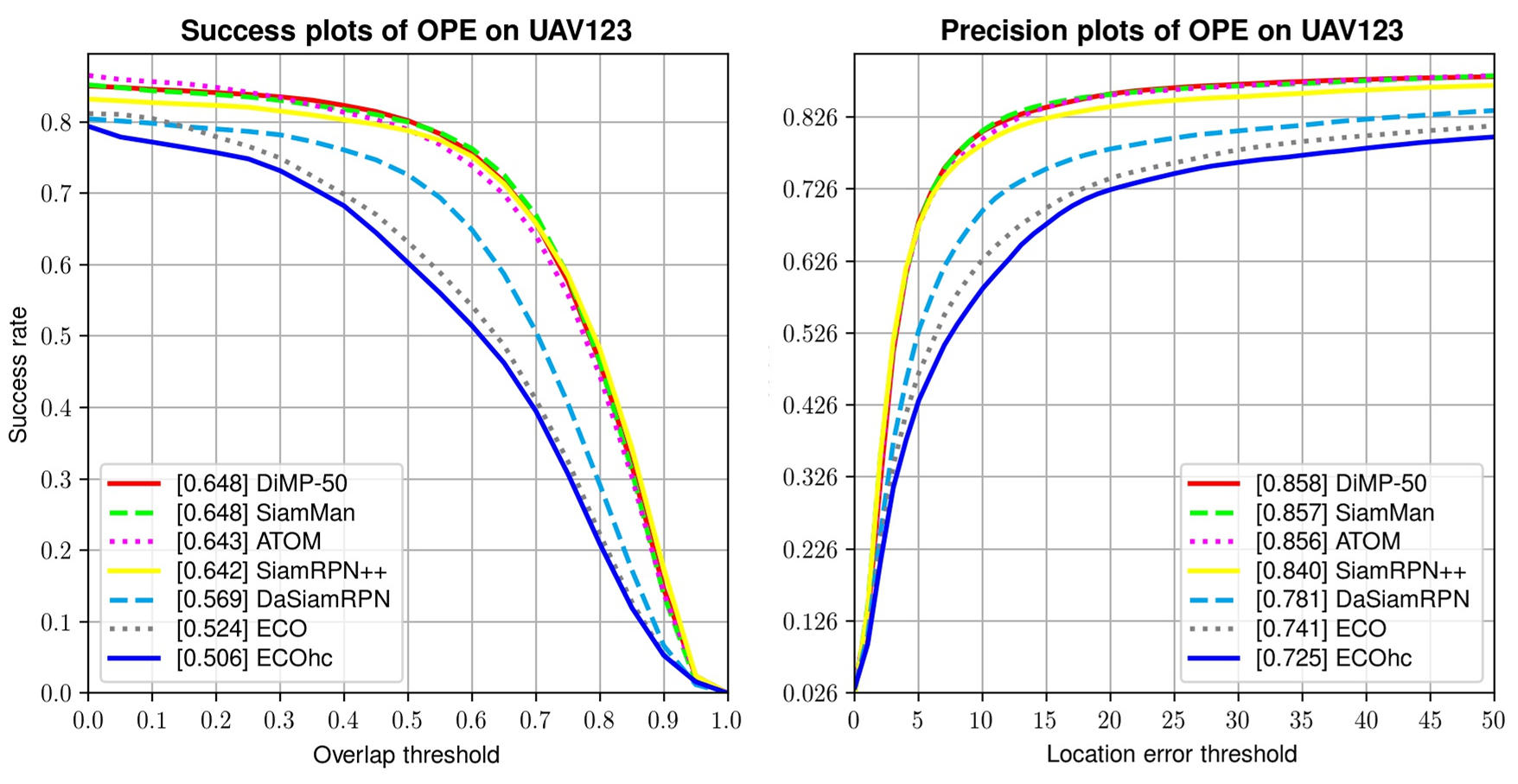}
\caption{Success and precision plots on the UAV123 dataset \cite{DBLP:conf/eccv/MuellerSG16}.}
\label{fig:uav}
\end{figure}

{\flushleft {\bf UAV123}.}
We also evaluate our SiamMan method on the UAV123 dataset \cite{DBLP:conf/eccv/MuellerSG16} in Figure \ref{fig:uav}. The dataset is collected from an aerial viewpoint, which includes $123$ sequences in total with more than $110,000$ frames. As shown in Figure \ref{fig:uav}, our method performs on par with the state-of-the-art tracker DiMP-50 \cite{DBLP:journals/corr/abs-1904-07220}, \ie, it produces the same success score $0.648$ but a little bit worse precision score ($0.857$ {\em vs.} $0.858$). Compared to SiamRPN++ \cite{DBLP:journals/corr/abs-1812-11703}, our method achieves higher success ($0.648$ {\em vs.} $0.642$) and precision scores ($0.857$ {\em vs.} $0.840$). It is attributed to the localization branch and the multi-scale attention module introduced in our tracker.

\begin{figure}[t]
\centering
\includegraphics[width=\linewidth]{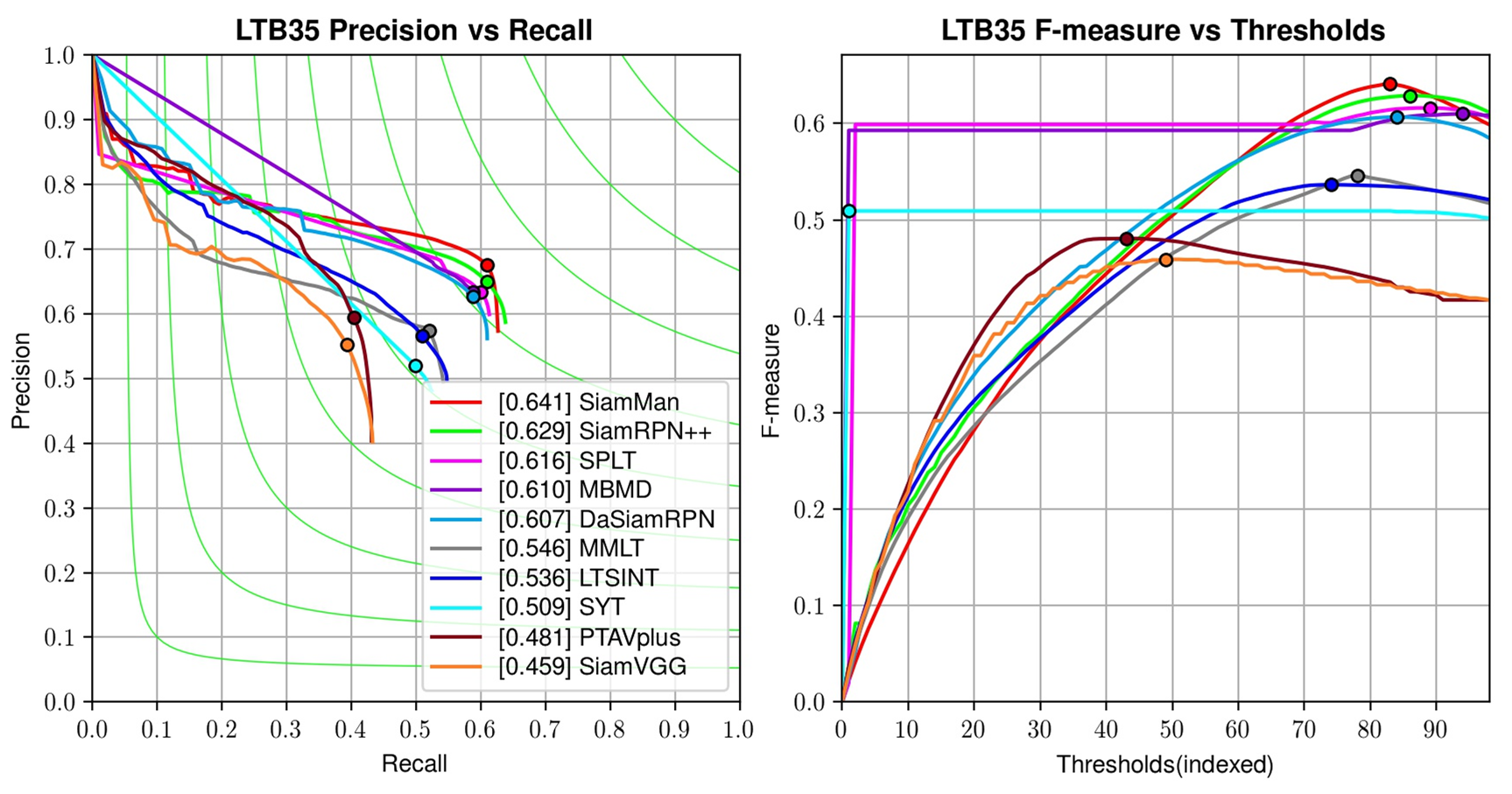}
\caption{Evaluation results on the LTB35 dataset \cite{DBLP:journals/corr/abs-1804-07056}, including recall and precision (left) and $\mathrm{F}$-score (right).}
\label{fig:ltb}
\end{figure}

{\flushleft {\bf LTB35}.}
In addition, we also evaluate our SiamMan tracker on the LTB35 dataset \cite{DBLP:journals/corr/abs-1804-07056}, which is first presented in the VOT2018-LT challenge \cite{DBLP:conf/eccv/KristanLMFPZVBL18}. It includes $35$ sequences with $14,687$ frames. Each sequence contains $12$ long-term target disappearing cases in average. We compare the proposed SiamMan method to several best-performing methods in the VOT2018-LT challenge \cite{DBLP:conf/eccv/KristanLMFPZVBL18} and SiamRPN++ \cite{DBLP:journals/corr/abs-1812-11703} in Figure \ref{fig:ltb}. As shown in Figure \ref{fig:ltb}, our method performs better than all those methods. Specifically, we achieve $64.1\%$ $\mathrm{F}$-score, \ie, $1.2\%$ higher than the second best method SiamRPN++ \cite{DBLP:journals/corr/abs-1812-11703} ($0.641$ {\em vs.} $0.629$). The results indicate that our method using the localization branch and multi-scale attention module performs well in long-term tracking even without using the re-detection module.

\subsection{Ablation Study}
To validate the effectiveness of different components, \ie, the localization branch, the global context module, and the multi-scale attention module, in our method, we conduct several ablation experiments on the challenging VOT2016 \cite{DBLP:conf/eccv/KristanLMFPCVHL16} and VOT2018 \cite{DBLP:conf/eccv/KristanLMFPZVBL18} datasets. Notably, we use the same parameter settings and training data for a fair comparison. In addition, we also analyze the performance of the Siamese network based trackers in $11$ attributes on the OTB100 dataset \cite{DBLP:journals/pami/WuLY15} to further demonstrate the effectiveness of the proposed method.

{\flushleft {\bf Localization branch.}}
If we remove the localization branch, the global context module, and the multi-scale attention module, our SiamMan degenerates to the original SiamRPN++ method \cite{DBLP:journals/corr/abs-1812-11703}. After integrating the localization branch, EAO is improved from $0.464$ to $0.488$ on VOT2016 and $0.415$ to $0.432$ on VOT2018, respectively. This significant improvements demonstrate that the localization branch is critical for the tracking performance.

{\flushleft {\bf Global context module.}}
In addition, we use global context module in the localization branch to capture long-range dependency. To demonstrate the effectiveness of the global context model, we construct a variant, \ie, remove the global context module from our SiamMan tracker, and evaluate it on VOT2016 and VOT2018, shown in the fourth column in Table \ref{tab:ablation}. As shown in the fourth and fifth columns in Table \ref{tab:ablation}, if we remove the global context module, the EAO scores drop $0.009$ ($0.513$ {\em vs.} $0.504$) and $0.015$ ($0.462$ {\em vs.} $0.447$), respectively. The results indicate that the global context module in the localization branch noticeably improve the tracking accuracy.

{\flushleft {\bf Multi-scale attention.}}
Furthermore, to validate the effectiveness of the multi-scale attention module, we construct a variant, \ie, remove the multi-scale attention module from our SiamMan tracker, and evaluate it on VOT2016 and VOT2018, shown in the third columns in Table \ref{tab:ablation}. As shown in Table \ref{tab:ablation}, we find that the multi-scale attention module significantly improves the performance of the proposed tracker on both VOT2016 and VOT2018, \ie, improving $0.019$ ($0.513$ {\em vs.} $0.494$) and $0.016$ ($0.462$ {\em vs.} $0.446$) EAOs. The learnable multi-scale attention module constructs an optimal combinations of multi-scale features from the siamese feature extraction subnetwork, which is effective to guide the three branches, \ie, the classification, regression, and localization branches, to exploit discriminative features better performance.

\begin{table}[t]
	\centering
   \small \setlength{\tabcolsep}{4.0pt}
	\caption{Effectiveness of different components in the proposed method based on EAO.}
	\vspace{2mm}
    \label{tab:ablation}		
	\begin{tabular}{c|ccccccc}
	\hline
	Component &\multicolumn{5}{c}{SiamMan}  \\
	\hline
	localization branch? 	&           & \checkmark  & \checkmark & \checkmark   & \checkmark \\
	global context?  &           &           & \checkmark &             & \checkmark \\
	multi-scale attention? 	&           &           &           & \checkmark   & \checkmark \\
	\hline
	\hline
	VOT2016                 & 0.464 & 0.488 & 0.494 & 0.504 & \textbf{0.513} \\
    VOT2018                 & 0.415 & 0.432 & 0.446 & 0.447 & \textbf{0.462} \\
	\hline
	\end{tabular}
\end{table}

\begin{figure}[t]
\centering
\includegraphics[width=\linewidth]{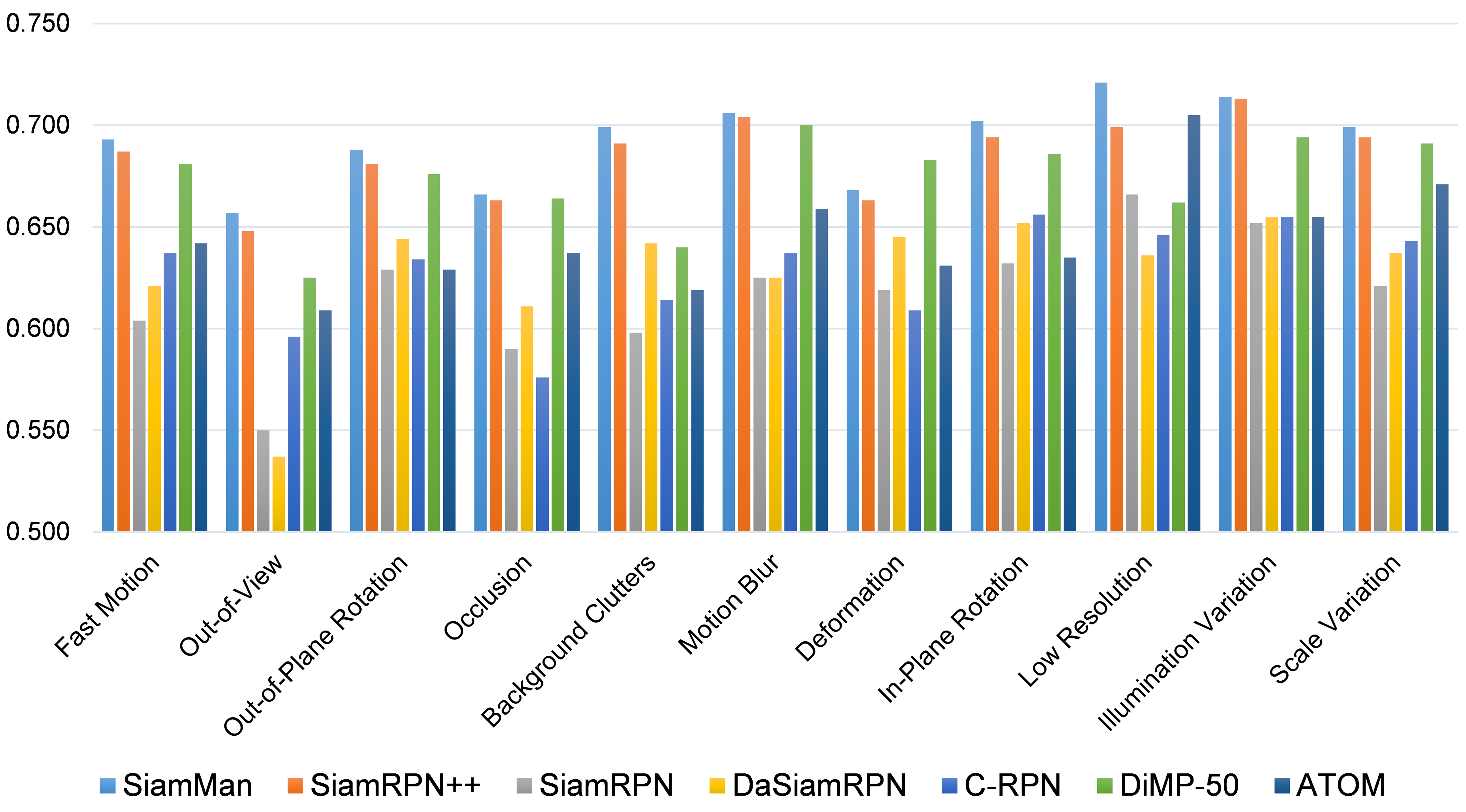}
\caption{Success score of the proposed method in each attribute on OTB100 \cite{DBLP:journals/pami/WuLY15}.}
\label{fig:attr}
\end{figure}

{\flushleft {\bf Performance on different attributes.}}
To verify the effectiveness of our method in detail, we also report the success score of the different trackers on different attributes in Figure \ref{fig:attr}. Compared to the Siamese network based trackers, \ie, SiamRPN++ \cite{DBLP:journals/corr/abs-1812-11703}, SiamRPN \cite{DBLP:conf/cvpr/LiYWZH18}, C-RPN \cite{DBLP:journals/corr/abs-1812-06148} and DaSiamRPN \cite{DBLP:conf/eccv/ZhuWLWYH18},  and other state-of-the-art methods, \ie, DiMP-50 \cite{DBLP:journals/corr/abs-1904-07220} and ATOM \cite{DBLP:journals/corr/abs-1811-07628}, our method performs the best in the most of the attributes, especially in {\em fast motion}, {\em out-of-view}, {\em low resolution} and {\em background clutters}. Most of the previous siamese network based trackers rely on the pre-set anchor boxes, making it difficult to adapt to different motion patterns and scales of targets, resulting in inaccurate tracking results in challenging scenarios such as {\em fast motion} or {\em low resolution} (\ie, indicating small scale target). The localization branch in the proposed method is effective to coarsely localize the target to help the regression branch to generate accurate results, making our tracker to be less sensitive to the variations of motion patterns and scales with the preset anchors.

\section{Conclusion}
In this paper, we propose a novel siamese motion-aware network for visual tracking, which integrates a new designed localization branch to deal with various motion patterns in complex scenarios. It coarsely localizes the target to help the regression branch to generate more accurate results, and lead to less tracking failures, especially when the fast motion, occlusion, and low resolution challenges occur. Moreover, we design a multi-scale attention module to guide these three branches to exploit discriminative features for better performance. Our tracker sets the new state-of-the-art on four challenging tracking datasets, \ie, VOT2016, VOT2018, OTB2015, and LTB35, and performs on par with the state-of-the-art on UAV123, with the real-time running speed $45$ frame-per-second.

{\small
\bibliographystyle{ieee_fullname}
\bibliography{references}
}

\end{document}